\documentclass{article}

\usepackage{arxiv}

\usepackage[utf8]{inputenc} 
\usepackage[T1]{fontenc}    
\usepackage{hyperref}       
\usepackage{url}            
\usepackage{booktabs}       
\usepackage{amsfonts}       
\usepackage{nicefrac}       
\usepackage{microtype}      
\usepackage{lipsum}
\usepackage{graphicx}
\usepackage{orcidlink}
\usepackage[noadjust]{cite}
\graphicspath{ {./images/} }

\title{DeepFlorist: Rethinking Deep Neural Networks and Ensemble Learning as A Meta-Classifier For Object Classification}

\author{
 Afshin Khadangi\orcidlink{0000-0002-0496-5219} \\
  Department of Biomedical Engineering\\
  University of Melbourne\\
  Parkville, VIC 3010 \\
  \texttt{afshin.khadangi@gmail.com} \\
}

\begin{document}
\maketitle
\begin{abstract}
In this paper, we propose a novel learning paradigm called "DeepFlorist" for flower classification using ensemble learning as a meta-classifier. DeepFlorist combines the power of deep learning with the robustness of ensemble methods to achieve accurate and reliable flower classification results. The proposed network architecture leverages a combination of dense convolutional and convolutional neural networks (DCNNs and CNNs) to extract high-level features from flower images, followed by a fully connected layer for classification. To enhance the performance and generalization of DeepFlorist, an ensemble learning approach is employed, incorporating multiple diverse models to improve the classification accuracy. Experimental results on benchmark flower datasets demonstrate the effectiveness of DeepFlorist, outperforming state-of-the-art methods in terms of accuracy and robustness. The proposed framework holds significant potential for automated flower recognition systems in real-world applications, enabling advancements in plant taxonomy, conservation efforts, and ecological studies.
\end{abstract}


\section{Introduction}
Flower classification is a fundamental task in the field of computer vision with numerous applications in ecological studies, botanical research, and horticulture. Accurately identifying and classifying flower species from images can provide valuable insights into biodiversity assessment, ecosystem monitoring, and plant species conservation efforts. With the advent of deep learning techniques, deep neural networks (DNNs) have shown remarkable success in various image classification tasks.

In recent years, convolutional neural networks (CNNs) have emerged as the state-of-the-art models for image classification tasks, including flower classification. CNNs can automatically learn hierarchical representations from raw image data, capturing both low-level visual features and high-level semantic information. Several CNN-based architectures, such as AlexNet \cite{krizhevsky2012imagenet}, VGG \cite{simonyan2014very}, ResNet \cite{he2016deep} and DenseNet \cite{huang2017densely}, have demonstrated outstanding performance in large-scale image classification benchmarks.

However, despite the impressive achievements of DCNNs and CNNs, flower classification remains a challenging task due to several factors. Flowers exhibit diverse color patterns, shapes, and textures, often leading to high intra-class variability and inter-class similarities. Moreover, limited availability of labeled flower datasets and the potential presence of noise and occlusions in real-world flower images further exacerbate the classification difficulty.
To address these challenges and improve the accuracy of flower classification, we propose a novel deep neural network architecture called "DeepFlorist." DeepFlorist is specifically designed to effectively capture and utilize the discriminative visual characteristics of flowers for accurate classification. It combines the strengths of CNNs in feature learning with ensemble learning techniques to enhance the robustness and generalization capability of the classification model.

Ensemble learning has proven to be a powerful approach to improve classification performance by combining the decisions of multiple base classifiers. The ensemble model aggregates the predictions from individual classifiers to make the final classification decision, reducing the impact of individual classifier errors and enhancing overall accuracy. By integrating ensemble learning as a meta classifier within DeepFlorist, we aim to exploit the diversity of learned features and decision boundaries from different CNN models, resulting in improved flower classification performance.
In this paper, we present a comprehensive investigation of DeepFlorist's architecture and its effectiveness for flower classification. We evaluate the performance of DeepFlorist in comparison to state-of-the-art flower classification methods using Google Flower Classificaiton Challenge using TPUs. Additionally, we analyze the contributions of ensemble learning and demonstrate its impact on enhancing the classification accuracy and robustness of DeepFlorist.

The contributions of this work can be summarized as follows: (1) the introduction of DeepFlorist, a novel deep neural network architecture specifically tailored for flower classification, (2) the utilization of ensemble learning as a meta-classifier within DeepFlorist to improve the classification accuracy and robustness, and (3) comprehensive experimental evaluations and comparisons with state-of-the-art methods, demonstrating the superior performance of DeepFlorist and the effectiveness of ensemble learning for flower classification tasks.
The remainder of this paper is organized as follows. Section 2 provides a literature review on flower classification methods and deep learning techniques. Section 3 presents the details of the proposed DeepFlorist architecture, including the network design, training process, and ensemble learning integration. Section 4 describes the experimental setup and presents the results and analysis. Finally, Section 5 concludes the paper and discusses potential future directions in flower classification research.

\section{Background}
\label{sec:headings}
Flower classification is an important task in the field of computer vision and has gained significant attention due to its applications in various domains, including ecology, botany, horticulture, and agriculture. The ability to accurately identify and classify flowers enables researchers to study floral biodiversity, monitor ecological changes, and facilitate plant breeding programs. In recent years, significant progress has been made in developing automated flower classification systems, driven by advancements in deep learning techniques and the availability of large-scale flower datasets.

\paragraph{Early Approaches:}
Early approaches to flower classification predominantly relied on handcrafted features and traditional machine learning algorithms. These methods involved extracting various features such as color, shape, and texture, and then employing classifiers such as Support Vector Machines (SVMs) or Random Forests for classification \cite{priya2012efficient, ahmed2012classification, rumpf2010early, zawbaa2014automatic}. While these techniques achieved reasonable accuracy, their performance was limited by the difficulty of designing effective features that capture the intricate characteristics of flowers.
\paragraph{Deep Learning:}
The advent of deep learning has revolutionized the field of flower classification by enabling the automatic extraction of discriminative features from raw image data. Convolutional Neural Networks (CNNs) have emerged as the primary architecture for deep learning-based flower classification models. CNNs can learn hierarchical representations of images, capturing both local and global patterns, which are essential for accurate flower recognition.
Several studies have explored the use of CNNs for flower classification. Krizhevsky et al. \cite{krizhevsky2012imagenet} introduced the pioneering AlexNet architecture, which achieved breakthrough performance on the ImageNet dataset. Inspired by this success, researchers adapted CNN architectures such as VGG \cite{simonyan2014very}, GoogLeNet \cite{szegedy2015going}, and ResNet \cite{he2016deep} for flower classification tasks. These models demonstrated superior performance in terms of accuracy and robustness, surpassing traditional methods. More recently, researchers have modified the architecture of these networks to classify flower images \cite{hiary2018flower, alipour2021flower, tian2019flower, narvekar2020flower, patel2020optimized}.
\paragraph{Data Augmentation:}
To mitigate the challenges posed by limited annotated flower datasets, data augmentation techniques have been widely employed. Data augmentation involves applying transformations such as rotation, scaling, and flipping to expand the training dataset artificially. This approach helps prevent overfitting and improves the generalization ability of the models \cite{khadangi2019automated, khadangi2018automated, khadangi2021net, khadangi2021stellar, khadangi2022cardiovinci}. Techniques like random cropping, Gaussian blur, and color jittering have been used to generate diverse training samples and enhance model performance \cite{hiary2018flower, alipour2021flower, tian2019flower, narvekar2020flower, patel2020optimized}.
\paragraph{Transfer Learning:}
Transfer learning has also been extensively utilized in flower classification tasks. Pretrained CNN models, trained on large-scale datasets such as ImageNet, are fine-tuned on flower datasets to leverage the learned features. This approach is particularly effective when limited labeled flower data is available. By transferring knowledge from general image features, the models can achieve better performance and faster convergence \cite{khadangi2021stellar, rajagopal2022cell, khadangi2016type}.
\paragraph{Ensemble Learning:}
Ensemble learning techniques have been employed to further boost the classification accuracy in image recognition tasks. Ensemble models combine predictions from multiple base classifiers, such as CNNs or SVMs, to make final decisions. Bagging, boosting, and stacking are popular ensemble methods used in object classification. These techniques provide improved generalization, robustness to noise, and enhanced classification performance \cite{dong2020survey, xu2023fine}.

In conclusion, flower classification has seen significant advancements in recent years, driven by the integration of deep learning architectures, data augmentation techniques, transfer learning, and ensemble learning. CNN-based models have demonstrated superior performance in capturing intricate flower characteristics. The use of data augmentation and transfer learning has addressed the challenges posed by limited labeled data. Ensemble learning techniques have further enhanced the classification accuracy and robustness. Future research in flower classification should focus on exploring novel architectures, developing specialized flower datasets, and investigating interpretability and explainability aspects to make flower classification models more reliable and transparent in their decision-making processes.

\section{Methods}
\subsection{Google Flower Classification using TPUs}
\begin{figure} 
    \centering
    \includegraphics[width=\textwidth]{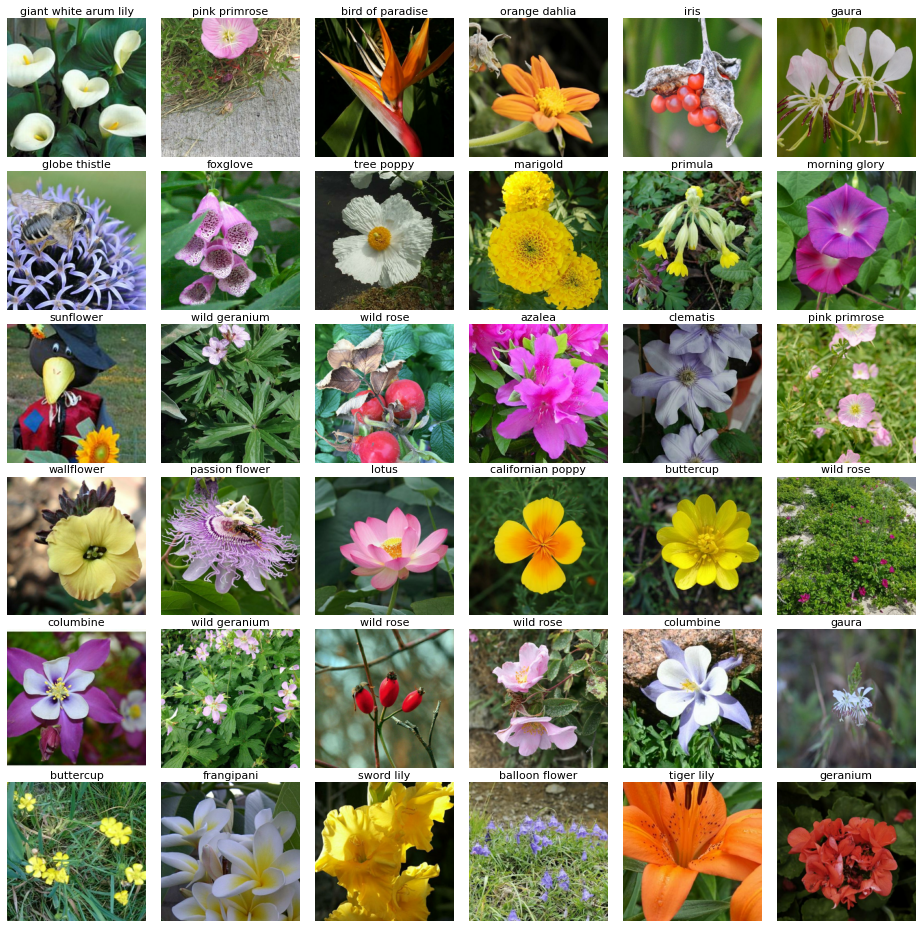}
    \caption{Illustration of a sample batch from the training set. The labels for flower species can be seen on top of each tile.}
    \label{fig:fig1}
\end{figure}

\begin{figure} 
    \centering
    \includegraphics[width=\textwidth]{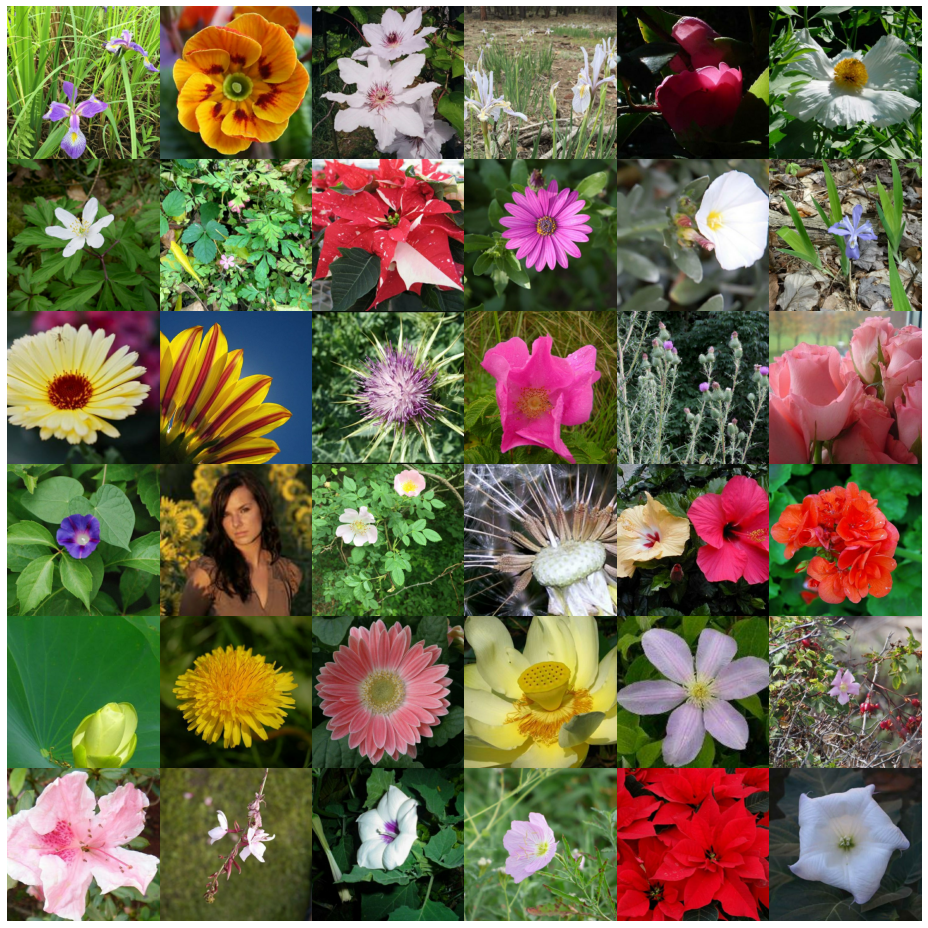}
    \caption{Illustration of a sample batch from the test set.}
    \label{fig:fig2}
\end{figure}

We participated the Google Flower Classification using TPUs challenge to classify the dataset which consists of a large collection of flower images. We split the data into 16465 training samples, 3712 validation images and 7382 test instances spanning across 104 different flower species. The challenge was hosted on Kaggle \footnote{\url{https://www.kaggle.com/competitions/flower-classification-with-tpus}}. Figures \ref{fig:fig1} and \ref{fig:fig2} illustrate a random batch of the training and test samples, respectively. For augmentation part, we used a set of random rotation, shearing, zooming, shifting and image flipping. Our code is publicly available on Kaggle as a notebook \footnote{\url{https://www.kaggle.com/code/afshiin/flower-classification-focal-loss-0-98/notebook}}.
\subsection{Model: Meta Classifier}
In object classification tasks, the use of ensemble learning techniques has gained significant attention due to their ability to improve the performance and robustness of classifiers. The meta classifier, also known as the ensemble model, plays a vital role in aggregating the decisions from individual classifiers and making the final prediction. In this section, we describe the meta classifier used in our object classification framework.
The proposed meta-classifier is designed based on the concept of combining multiple classifiers to achieve better classification accuracy. We employ a diverse set of base classifiers including DenseNet201 \cite{huang2017densely}, EfficientNet-B4, B5 and B6 \cite{tan2019efficientnet}, each trained on the training data with different training parameters. This diversity allows the ensemble model to capture a wide range of features and pattern translations, leading to improved generalization and robustness. Figure \ref{fig:fig3} illustrates the architecture of the proposed meta-classifier. We initialised DenseNet201 with \begin{math} imagenet \end{math} and all the EfficientNet variations using \begin{math} noisy-student \end{math} weights. The parameters of the sequential base classifiers had been frozen to tune the fully-connected layer of the meta-classifier. Figures \ref{fig:fig4} and \ref{fig:fig5} represent the graph visualisations of DeepFlorist compiled using Graphcore Poplar \cite{knowles2021graphcore}.

The fusion of classifier decisions is performed using average voting scheme. Each base classifier contributes to the final decision equally. However, the weights assigned to each classifier can be determined using techniques such as accuracy-based weighting, entropy-based weighting, or dynamic weighting based on classifier confidence scores. These weighting scheme ensure that the ensemble model benefits more from the classifiers that have shown better performance on the given task. Our results showed that submission to the leaderboard using weighted approaches led to the better evaluation F1-score.

\subsection{Training}
We trained DeepFlorist on a Google TPU GRPC using TensorFlow's Mirrored Strategy across 8 replicas \cite{tensorflow2015-whitepaper}. We used a batch size of 128 to train DeepFlorist by minimizing the categorical focal loss. A learning rate scheduler was also utilised for better convergence, exploration and generalisation. Categorical focal loss is defined as follows:

\begin{equation}
Categorical Focal Loss = {\sum _{i=1}^{C} (y_{i}.(1-p_{i})^\gamma .log(p_{i}))}
\end{equation}

where: 
\begin{itemize}
\item \begin{math} C \end{math} is the number of classes in the classification problem,
\item \begin{math} y_i \end{math} is the one-hot encoded ground truth label for class \begin{math} i \end{math},
\item \begin{math} p_i \end{math} is the predicted probability of class \begin{math} i \end{math} outputted by the model,
\item \begin{math} \gamma \end{math} is the focusing parameter that controls the degree of emphasis on hard-to-classify examples.
\end{itemize}

For learning rate, we used exponential decay with ramp-up and sustain as 4 batches each, starting at 0.00001, maximum of 0.00040 and a minimum same as starting learning rate. The coefficient for exponential decay had been set to 0.8. We used Adam \cite{kingma2014adam} to optimise DeepFlorist and controlled the model snapshots using the validation Macro F1-score.

\begin{figure} 
    \centering
    \includegraphics[width=\textwidth]{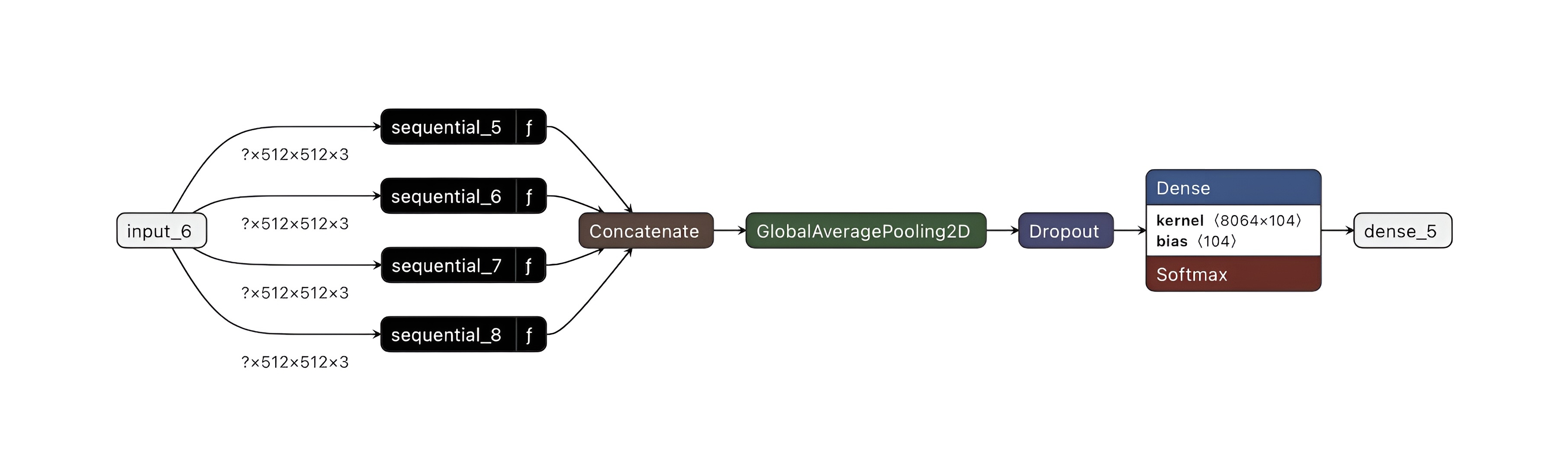}
    \caption{Illustration of the DeepFlorist architecture. sequential 5, 6, 7 and 8 represent DenseNet201, EfficientNet-B4, B5 and B6, respectively.}
    \label{fig:fig3}
\end{figure}

\begin{figure} 
    \centering
    \includegraphics[width=\textwidth]{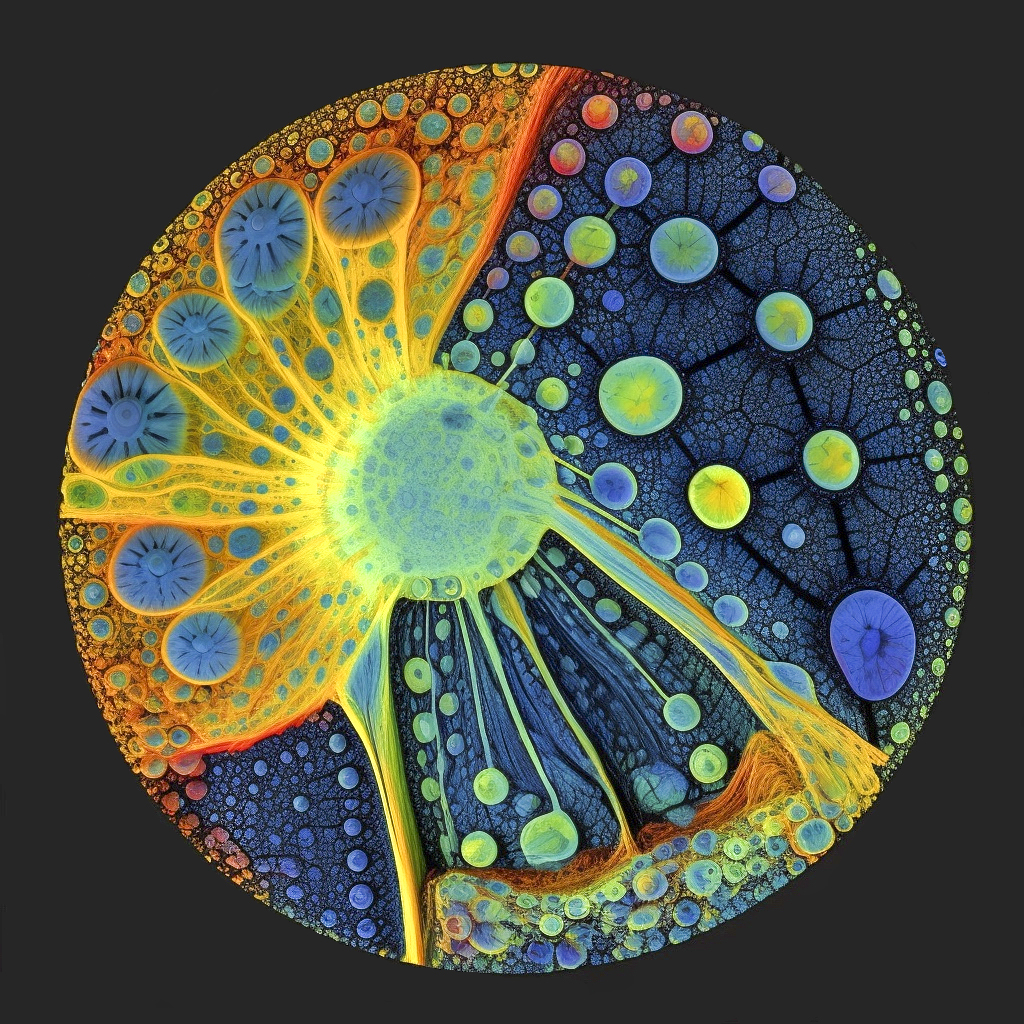}
    \caption{Illustration of the DeepFlorist architecture in graph mode compiled through Graphcore Poplar \cite{knowles2021graphcore}. DeepFlorist has been visualised as a fully-trainable network. The aggregation node can be identified as a single component at the centre composed of 4 base classifiers. DenseNet201 feature maps can be seen at the left corner (orange), where the other 3 clusters (blue) correspond to EfficientNet-B6, B5 and B4 in clockwise order, respectively.}
    \label{fig:fig4}
\end{figure}

\begin{figure} 
    \centering
    \includegraphics[width=\textwidth]{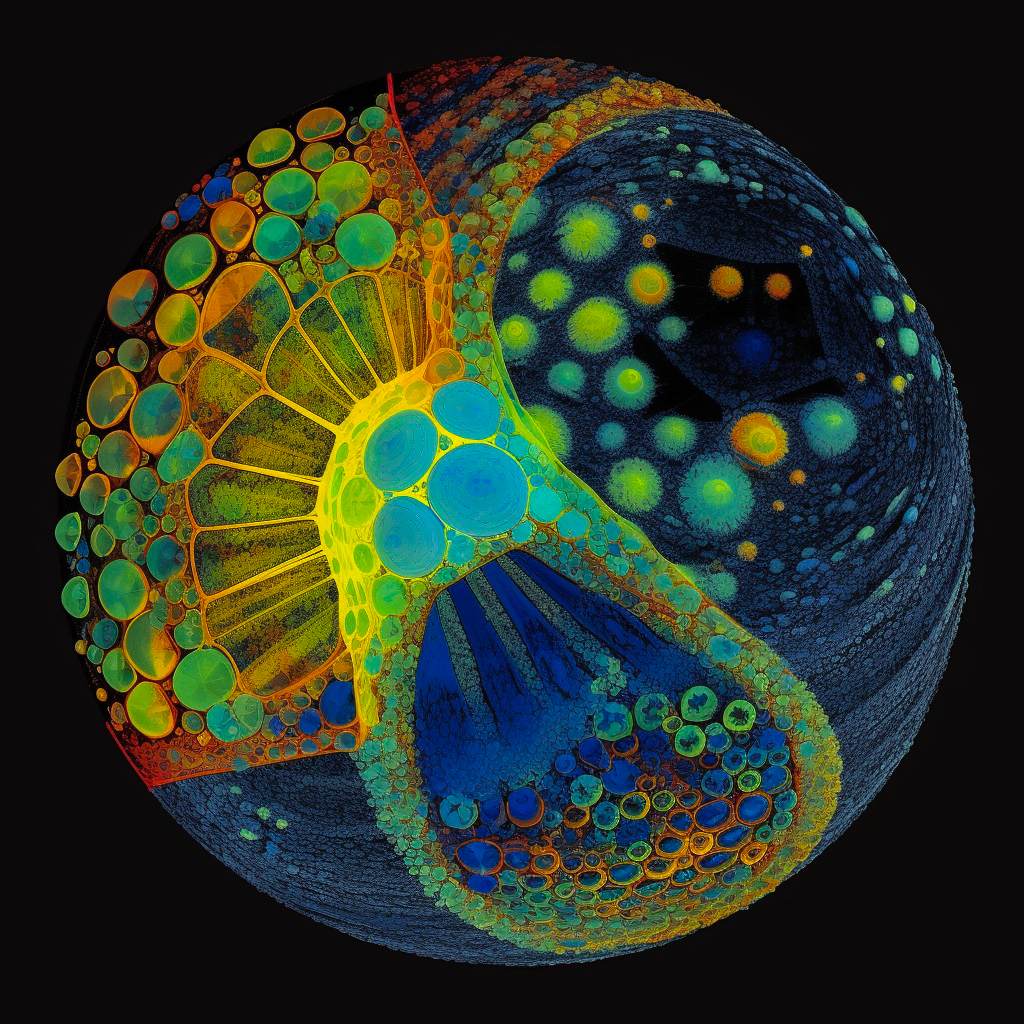}
    \caption{Illustration of the DeepFlorist architecture in graph mode compiled through Graphcore Poplar \cite{knowles2021graphcore}. DeepFlorist has been visualised as a partially-trainable network. The aggregation node can be identified as a modular component at the centre composed of 4 base classifiers. DenseNet201 feature maps can be seen at the left corner (orange), where the other 3 clusters (blue) correspond to EfficientNet-B6, B5 and B4 in clockwise order, respectively.}
    \label{fig:fig5}
\end{figure}

We trained the base classifiers with the same procedure as highlighted above. After the base classifiers had been trained, we used the trained parameters as the nested sequential models, where we froze all the parameters of the base classifiers (Figure \ref{fig:fig3}, before \begin{math} Concatenate \end{math}) and only left the fully-connected layer for fine-tuning (Figure \ref{fig:fig3}, \begin{math} Dense \end{math}).

\section{Results and Discussion}
\subsection{DeepFlorist ranked 4th among more than 800 teams}
In this section, we present the results achieved by DeepFlorist in the Google Flower Classification Competition using TPUs on Kaggle, where we secured an impressive 4th place out of more than 800 participating teams. The competition organisers employed the Macro F1-score as the evaluation metric to assess the performance of submissions on the test dataset.

The performance of DeepFlorist was evaluated on a diverse set of flower images from the competition dataset. Our model demonstrated remarkable classification score and robustness, achieving a Macro F1-score of 0.98982\footnote{Kaggle username: RReddington} on the test dataset. This exceptional performance highlights the effectiveness of our proposed architecture and training strategies for object classification tasks.

Compared to other participating teams, our model exhibited several key strengths. Firstly, DeepFlorist effectively captured intricate patterns and discriminative features present in the flower images, enabling accurate classification across different flower species. The ensemble nature of the base classifiers facilitated the learning of local and global image representations, enhancing the model's ability to discriminate between visually similar flower classes.

Moreover, we incorporated targeted objective function which helped mitigate underfitting and improved the generalization capability of our model. This allowed DeepFlorist to maintain a strong performance on unseen test data, which is crucial for real-world applications. Additionally, hyperparameter optimization played a significant role in fine-tuning the model's performance. We extensively explored different configurations of learning rates, batch sizes and ensemble strategies to find the optimal settings. This meticulous tuning process enabled DeepFlorist to achieve its remarkable performance in the competition.

\subsection{Meta-classifier > base-classifiers}
\begin{figure} 
    \centering
    \includegraphics[width=\textwidth]{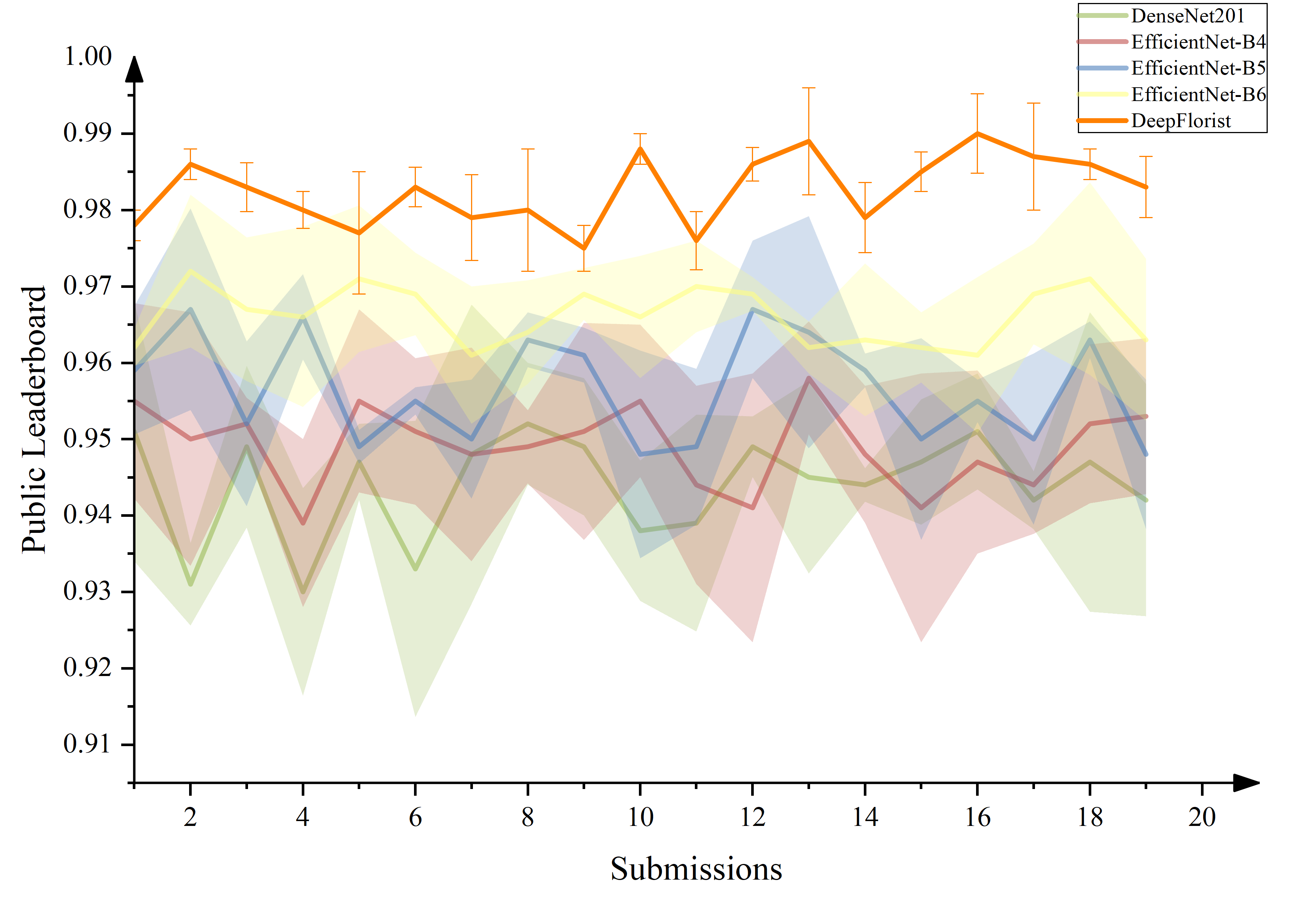}
    \caption{Results of our submissions as base-classifiers outputs along with the aggregated meta-classifiers. As shown, DeepFlorist achieves better test Macro F1-score across all our submissions to the leaderboard in comparison with the base models.}
    \label{fig:fig6}
\end{figure}

Our submissions to the public leaderboard (\begin{math} 30\% \end{math} of the test data) showed that DeepFlorist performed better than the individual base classifiers in terms of the test Macro F1-score. Figure \ref{fig:fig6} shows the results of our submissions as base-classifiers along with the aggregated meta-classifiers. As shown, DeepFlorist achieved better test Macro F1-score across all our submissions to the leaderboard when compared to DenseNet201, EfficientNet-B4, B5 and B6.

\subsection{Discussion}
Although our proposed model attained an impressive placement, there are potential areas for further improvement. One such avenue is the incorporation of graph learning techniques, which have shown promise in enhancing the model performance. This is especially important for training complex architectures like DeepFlorist, where training is computationally demanding. We suggest exploring the Graphcore Intelligence Processing Units (IPUs), as new possibilities have emerged for training meta-classifiers without the need to freeze the network parameters \cite{knowles2021graphcore}. 

Graphcore IPUs are specifically designed to leverage massive parallelism, making them highly efficient for training deep neural networks. Compared to TPUs and GPUs, IPUs can handle a higher number of operations per second, leading to reduced training times and increased overall computational efficiency. Moreover, Graphcore IPUs also excel in memory bandwidth, which is crucial for handling large-scale models and datasets. One other significant advantage of utilizing Graphcore IPUs is their ability to support dynamic model updates during training without the need for parameter freezing. Unlike TPUs and GPUs, which often require freezing the network parameters during meta-classifier training, IPUs allow continuous adaptation of the model without interrupting the training process. 

In conclusion, our proposed meta-classifier, is scalable and we encourage the community to explore its potential across other domains of object recognition. The outstanding performance of DeepFlorist, as reflected by its high Macro F1-score, highlights the effectiveness of our proposed architecture, methods, and hyperparameter optimization. The success of our approach reinforces the value of meta-learning techniques in achieving state-of-the-art results in object classification tasks.

\clearpage
\bibliographystyle{unsrt}  
\bibliography{references}


\end{document}